# Breast Cancer Detection and Diagnosis: A comparative study of state-of-the-arts deep learning architectures

Brennon Maistry[1] and Absalom E. Ezugwu[2][0000-0002-3721-3400]

[1] School of Mathematics, Statistics, and Computer Science, University of KwaZulu-Natal, King Edward Avenue, Pietermaritzburg Campus, Pietermaritzburg, 3201, KwaZulu-Natal, South Africa

[2] Unit for Data Science and Computing, North-West University, 11 Hoffman Street, Potchefstroom, 2520, South Africa
`Absalom.Ezugwu@nwu.ac.za`

**Abstract.** Breast cancer is a prevalent form of cancer among women, with over 1.5 million women being diagnosed each year. Unfortunately, the survival rates for breast cancer patients in certain third-world countries, like South Africa, are alarmingly low, with only 40% of diagnosed patients surviving beyond five years. The inadequate availability of resources, including qualified pathologists, delayed diagnoses, and ineffective therapy planning, contribute to this low survival rate. To address this pressing issue, medical specialists and researchers have turned to domain-specific AI approaches, specifically deep learning models, to develop end-to-end solutions that can be integrated into computer-aided diagnosis (CAD) systems. By improving the workflow of pathologists, these AI models have the potential to enhance the detection and diagnosis of breast cancer. This research focuses on evaluating the performance of various cutting-edge convolutional neural network (CNN) architectures in comparison to a relatively new model called the Vision Transformer (ViT). The objective is to determine the superiority of these models in terms of their accuracy and effectiveness. The experimental results reveal that the ViT models outperform the other selected state-of-the-art CNN architectures, achieving an impressive accuracy rate of 95.15%. This study signifies a significant advancement in the field, as it explores the utilization of data augmentation and other relevant preprocessing techniques in conjunction with deep learning models for the detection and diagnosis of breast cancer using datasets of Breast Cancer Histopathological Image Classification.

## 1 Introduction

Breast cancer encompasses all cancers found in the breast, primarily in the epithelium of the ducts or lobules of the glandular tissue [1]. It is a metastatic disease that frequently spreads to other organs, such as the lungs, liver, and spine [2]. Unfortunately, breast cancer is often incurable. In 2020, the World Health Organization (WHO) reported 2.3 million women diagnosed with breast cancer worldwide, resulting in 685,000 deaths [1]. Additionally, it is projected that in 2022, the United States will witness 43,780 deaths from breast cancer, affecting 43,250 women and 530 men [3]. With over 1.5 million women diagnosed annually across the globe [2], breast cancer remains one of the most prevalent cancers among women.

The treatment of breast cancer has shown high efficacy, boasting a five-year survival rate of 90% following diagnosis. However, this favorable probability is primarily achievable in higher-income countries like the United States [1, 3]. In lower-income nations such as India and South Africa, the five-year survival rates plummet dramatically to 66% and 40% respectively [1]. This rapid decline can be attributed to limited resources, high costs of medical examinations, and extended waiting times for access to pathologists' services. Breast pathology analysis and cancer treatment decisions often rely on the skill level and experience of the pathologist, making the workflow tedious and subjective [4].

Histopathology, as defined by Robertson et al. [4], involves the examination of tissue specimens after fixation in formalin, paraffin embedding, and mounting thin histologic sections onto glass slides [4]. While mammography serves as the most popular non-invasive clinical screening technique for breast cancer diagnosis, it faces challenges such as reduced sensitivity influenced by breast density and limited effectiveness in women under the age of 40 [5]. Hence, histopathological image analysis remains the gold standard for breast cancer diagnosis and serves as the medical imagery used in this research.

The introduction of whole-slide digital image scanners has transformed the pathologist's workflow into a digital process, enhancing the convenience of patient diagnostic analysis. Furthermore, numerous companies now harness the power of artificial intelligence (AI) to develop machine learning solutions that optimize the pathologists' workflow. IBM Watson Health, for instance, offers the "IBM Imaging Workflow Orchestrator with Watson," a cloud-based Software-as-a-Service (SaaS) program utilizing AI to assist radiologists in analyzing patient images and extracting associated dataset features [6]. However, a limitation of machine learning methods in the medical domain is the meticulous annotation required, which remains subjective to the annotator's domain knowledge.

To optimize the pathologist's workflow using Computer-Aided Diagnosis (CAD), researchers have started exploring Deep Learning (DL) to overcome challenges encountered with machine learning models. Deep Learning, a subset of machine learning, emulates the learning process of the human brain [7]. By learning from examples, deep learning can extract necessary features based on input and output target classes. The independence of dependencies required by DL models validates their significance in constructing fully automated CAD systems.

Convolutional Neural Networks (CNNs) have shown remarkable performance in breast cancer binary classification and feature extraction. In a study by researchers [8], a fine-tuned Inception-v3 model was employed to classify biopsy images as either benign or malignant tumors. This approach achieved an accuracy surpassing state-of-the-art models at various levels of image magnification. Another study [9] utilized a pretrained ResNet50 CNN model for feature extraction and a Logistic Regression classifier to determine the malignancy of histopathological tumor images, achieving an accuracy of 93.27%. These results demonstrate the potential of CNNs for breast cancer diagnosis. However, researchers [10] raised a significant concern regarding the limited availability of medical data, which hampers the effective training of more complex CNN models like ResNet50. Moreover, shallower CNN models, such as AlexNet, tend to overfit smaller datasets.

Although the Transformer architecture has gained prominence in Natural Language Processing (NLP), in 2020, Google researchers adopted the Transformer framework for computer vision, resulting in the Vision Transformer (ViT) model [11]. Initial findings indicate that ViT surpasses the state-of-the-art CNN models of that time. This groundbreaking research necessitates a comparison between the traditional Transformer architecture and its CNN counterparts.

As advocated by authors [10], the size of the dataset significantly influences the effectiveness of CNN models used. Therefore, to develop effective Computer-Aided Diagnosis (CAD) systems, it is crucial to select the most suitable CNN architectures. Consequently, a comparative study was conducted to assess the performance of different CNN architectures in medical imagery and compare them to the recently introduced ViT model. This paper aims to employ deep learning techniques to automate breast cancer detection using histopathological images. Furthermore, it seeks to compare and evaluate the performance of AlexNet, Inception-v3, and ResNet18 CNN architectures against the ViT architecture.

The technical contributions of this paper are as follows:
i. Proposal of enhanced deep learning models using data augmentation techniques such as rotation and flipping, and k-fold cross-validation to assess the predictive capabilities of CNN models on unseen data.
ii. Comparison of the performance of AlexNet, Inception-v3, and ResNet18 CNN architectures against the ViT architecture.

The rest of this paper is organized as follows: Section 2 presents a literature review, highlighting previous work in this domain and identifying their limitations, thereby justifying the importance of this study. Section 3 outlines the methods and techniques employed to address the problem at hand. Section 4 discusses the experimental results and compares the models implemented in this paper with the findings from existing literature. Finally, Section 5 concludes the paper and suggests future extensions or improvements to this study.

## 2       Literature Review

A considerable amount of research has been conducted on CNN architectures for breast cancer detection, which has further extended to hybrid models incorporating machine learning classifiers and ensemble learning techniques to enhance classification accuracy. On the other hand, the utilization of ViT models in the medical domain is still in its early stages and not as extensively explored as CNN architectures in this field. This section provides an overview of some key techniques and findings in the existing literature.

In their work, Senan et al. [12] proposed a system for detecting breast cancer using histopathological images by combining CNN with hierarchal voting applications. Their objective was to accurately classify images as benign or malignant tumors. The researchers employed the widely used BreakHis dataset [13] and implemented data augmentation and normalization techniques to address the challenges posed by limited training data for DL models. Transfer learning was applied to a CNN based on the AlexNet architecture, where the original classification layer (designed for classifying 1000 classes) was replaced with a binary classification-capable fully connected layer. The classification of images involved both patch-level and image-level voting applications. While this system achieved favorable results, with an accuracy of 95% and an Area Under Curve (AUC) of 99.36%, further investigation can be conducted by comparing the system against a standard CNN model using the AlexNet architecture, trained and tested on the data prepared by the proposed system. This analysis would provide insights into whether the performance improvements were due to the enhanced dataset preparation, the hierarchical voting applications, or a combination of both. Additionally, cross-validation can be employed to validate the model's performance more rigorously.

In the study conducted by Achuthan et al. [14], the performance of the AlexNet architecture was compared with the VGG16 and VGG19 CNN architectures. All models were pre-trained, and a 5-fold cross-validation approach was employed. The results of the comparison indicated that the AlexNet model outperformed the other architectures, emerging as the best-performing model. The researchers further evaluated the AlexNet model in conjunction with different machine learning classifiers, including Decision Tree, Random Forest, and K-Nearest Neighbors. The model that yielded the highest accuracy (87.5%) utilized the AlexNet CNN for feature extraction combined with the Random Forest classifier, surpassing the standard AlexNet CNN by 1.5%. However, a limitation of this study was the relatively small dataset obtained from Kaggle, consisting of only 4000 images. Although data augmentation techniques were applied, the limited dataset size posed challenges for more complex models like VGG16 and VGG19, while the shallower nature of the AlexNet made it less prone to overfitting. To address this limitation, the authors suggest increasing the number of folds in cross-validation and implementing image normalization to account for the small dataset size.

In the research presented by Ranjan et al. [15], the AlexNet CNN was employed in conjunction with hierarchical CNN approaches for classification purposes. The researchers demonstrated the advantages of training the entire pre-trained AlexNet network rather than solely training the final connected layer. Training the full network resulted in a 6% improvement in accuracy, reaching 83% compared to 77% when training only the fully connected network. The classification approach explored two strategies. In the first approach, the AlexNet CNN determined whether an image was classified as "Normal" or "Rest," and if classified as "Rest," another CNN determined the specific class among "Invasive," "Insitu," or "Benign." The second approach utilized the AlexNet classifier to classify an image as "Normal" or "Rest," followed by the application of majority voting for three binary CNN models. This approach yielded better results, achieving an accuracy of 95%. Similar to the previous work, the limitation identified in this study was the dataset size. The researchers utilized the BACH Challenge 2018 dataset, which contained only 400 images. Additionally, no data augmentation was employed, highlighting the need for cross-validation to further validate the system's performance and the use of some form of data augmentation to reduce generalization error.

In the study conducted by Benhammou et al. [8], preliminary performance results of a pre-trained Inception-v3 CNN model were presented. However, no data augmentation or preprocessing was applied to the BreakHis dataset, making it a preliminary result. The study also explored the DeCaf approach to transfer learning and used the Inception-v3 model for binary classification, classifying the images as "malignant" or "benign." The results showed the potential of Inception-v3 as an effective model, achieving an accuracy of 90.2% on images with 40x magnification. However, CNN models that were fed preprocessed and augmented data outperformed their approach, reaching an accuracy of 96.1%. This highlights the importance of preprocessing and data augmentation for achieving good performance metrics in a model.

In the work conducted by Xiang et al. [17], an Inception-v3 CNN model was fine-tuned and modified for binary classification on the BreakHis dataset. This study emphasized the significance of cross-validation and data augmentation. The Inception-v3 model trained without data augmentation achieved an accuracy of 92.8%, but when cross-validation and data augmentation were applied, the performance increased by 2.9%, resulting in the best accuracy of 95.7%. These findings validate the importance of data augmentation and cross-validation when training CNN models on relatively smaller datasets, such as those found in medical imagery.

In a recent study by Aljuaid et al. [18], the performance of pre-trained CNNs, including ResNet18, Inception-v3, and ShuffleNet, was compared on the BreakHis dataset. The dataset was split into 65% for training and 35% for testing. Median and Gaussian filters were utilized to remove noise from the images while preserving their original features. Data augmentation was also applied to reduce overfitting. All models were trained using the more recent, smaller ImageNet dataset and then subjected to transfer learning. The evaluation of the models revealed that the ResNet18 CNN model achieved the best results, with an accuracy of 99.7% for binary classification and an accuracy of 97.81% for multiclass classification. To further improve the study, producing generalized results for each model could provide a better understanding of how the model performs with unfamiliar data. This could be achieved by applying k-fold cross-validation and a normalization technique for the images to obtain a more generalized performance of the model.

A study conducted by Thomas et al. [19] compared the performance of state-of-the-art CNN models with a ViT model in binary classification on the BreakHis dataset. The researchers utilized computer vision techniques to preprocess the images, including Adaptive Histogram Equalization, Multiscale Retinex with Color Restoration, and Median Filtering. Data augmentation techniques such as horizontal flipping and resizing to 224x224 were also applied. The study employed cross-validation as part of its training strategy, and the model achieved an outstanding accuracy of 96.7%. To further enhance this research, using normalized images and specialized techniques for histopathological images could be explored. Additionally, comparing the ViT model against end-to-end CNNs for classification could provide insights into the performance advantages or disadvantages of using a ViT model.

Based on the reviewed literature, all models exhibited strengths and weaknesses based on the authors' data preprocessing and augmentation approaches. This paper aims to evaluate the highlighted DL models using standard preprocessing and augmentation techniques, while also implementing k-fold cross-validation. This comprehensive strategy allows for an effective evaluation of the DL models' performance in classifying histopathological images and accurately diagnosing breast cancer.

## 3      Methods and Techniques

This section outlines the methods and techniques employed in this research. The process begins with data collection, followed by preprocessing and augmentation, which includes rotation and flipping of the images. The augmented dataset is subsequently divided into 10 folds to facilitate training of the deep learning (DL) models using k-fold cross-validation. After training the models, they are evaluated using unseen images from the dataset. Figure 1 provides an overview of the methodology.

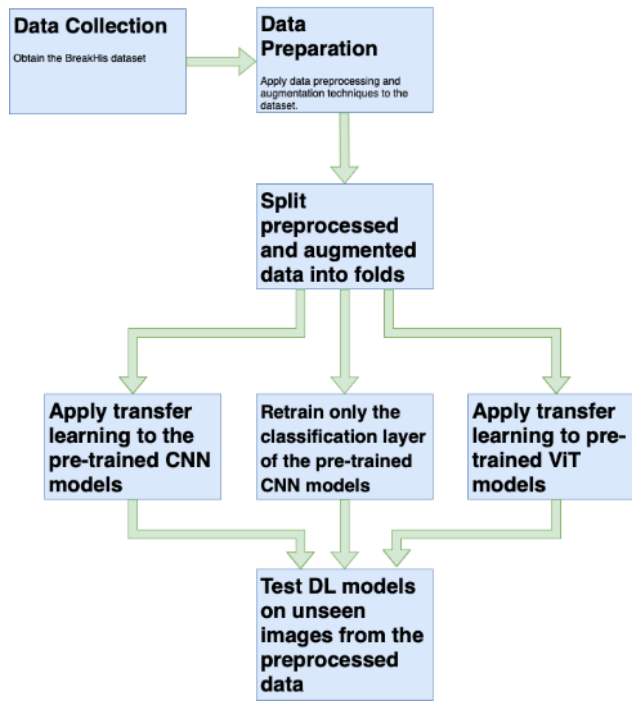

**Fig. 1.** Overview of the methodology

### 3.1 BreakHis Dataset

For this study, the widely recognized BreakHis dataset [13] was utilized. This dataset consists of a total of 7,909 images obtained from 82 patients at four different magnification levels (40x, 100x, 200x, and 400x). The dataset is categorized into two types: benign and malignant lesions. Among the images, 2,480 correspond to benign lesions, while the remaining 5,429 represent malignant lesions. All the images possess a resolution of 460x700 pixels. The BreakHis dataset was developed in collaboration with P&D laboratories and has emerged as the standard dataset for deep learning-based binary classification of breast cancer. Figure 2 displays an example image of a malignant lesion at various magnification levels, and Table 1 provides a more detailed breakdown of the dataset.

Table 1. Breakdown description of the BreakHis dataset.

| Magnification | Benign | Malignant | Total |
|---|---|---|---|
| 40x | 652 | 1370 | 1995 |
| 100x | 644 | 1437 | 2081 |
| 200x | 623 | 1390 | 2013 |
| 400x | 588 | 1232 | 1820 |
| Total | 2480 | 5429 | 7909 |

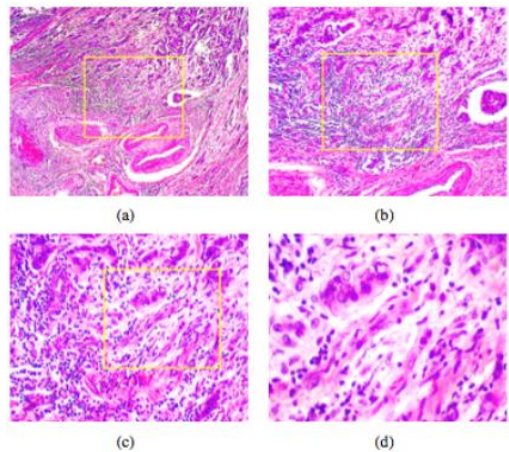

**Fig. 2.** Image of the malignant lesion at different magnification factors: (a) 40x, (b) 100X, (c) 200x, (d) 400x.

### 3.2 Data Preprocessing and Augmentation

The removal of noise is a crucial task in image processing as it contributes to the generation of higher-quality images. In the field of pathology, one persistent challenge faced by the industry is the inconsistency in slide preparation. This inconsistency arises from variations in stain manufacturers, staining procedures, and even differences in the storage time of slides, all of which have an impact on the digital representation of histopathological images. To address this issue, Macenko et al. [21] introduced an algorithm that effectively normalizes histopathological slides, enabling their utilization for quantitative analysis. Algorithm 1 presents the algorithm proposed by Macenko et al. [21].

**Algorithm 1:** Step for the normalization of histopathological images [21]

**Input:** RGB Slide
1. Convert RGB to OD
2. Remove data with OD intensity less than $\beta$
3. Calculate SVD on the OD tuples
4. Create plane from the SVD directions corresponding to the two largest singular values
5. Data onto the plane, and normalize to unit length
6. Calculate angle of each point wrt the first SVD direction
7. Find robust extremes ($\alpha$th and (100−$\alpha$)th percentiles) of the angle
8. Convert extreme values back to OD space

**Output:** Optimal Stain Vectors

In previous attempts to develop machine learning algorithms for breast cancer, the manual crafting of meticulous features was necessary. However, the advent of Convolutional Neural Network (CNN) models, with their convolutional layers, has enabled effective identification of key image features based on texture and histology structures. This autonomous feature extraction capability comes at the cost of CNN models being highly dependent on large amounts of data, which poses a challenge in many medical domains due to the limited availability of medical datasets compared to datasets like ImageNet.

To overcome the data scarcity issue, researchers have turned to data augmentation techniques. Data augmentation involves generating new images by applying transformations such as flipping, rotation, and cropping to existing ones. By employing this technique, researchers can significantly enhance the performance of deep learning models. This is supported by the findings of Zuluga-Gomez et al. [22], whose study demonstrated that effective data augmentation can result in a CNN model performing equally well as a model trained on a dataset that is 50% larger.

In this study, the images are first normalized using the stain normalization algorithm proposed by Macenko et al. [21]. This algorithm, specifically designed for histopathological slides/images, was chosen over traditional image processing techniques. The data augmentation technique employed follows the approach outlined by Krishevsky et al. [23]. The training images are resized using Bilinear Interpolation, based on the input dimension of the CNN models, and random crops are taken along with their horizontal flips. For testing and validation, five crops (the four corners and the center) and their horizontal flips are used. Similar computational procedures are implemented in this study, aligning with the approach of Krishevsky et al. [23]. The image transformation process is performed on the CPU using Python code, while the models are trained on the GPU. The data is transferred to the GPU only after being inserted into batches

### 3.3 CNN Architectures

For this research, the CNN models adopted include AlexNet, Inception-v3 and ResNet18.

**AlexNet Model.** The groundbreaking AlexNet model [23], introduced by Alex Krishevsky et al. in 2012, had a profound impact on the adoption of CNN models for computer vision tasks. One of the notable contributions of AlexNet was its ability to train on multiple GPUs, which revolutionized the field. Additionally, the model replaced traditional activation functions like the tanh function with Rectified Linear Units (ReLU), enabling faster training times compared to using tanh. The model employed GPU parallelization by distributing its kernels across multiple GPUs, with communication occurring only on specific kernels. Although this reduced training time in standard scenarios, Krishevsky et al. [23] mentioned that using GPU parallelization with cross-validation posed a challenge, requiring careful tuning of communication patterns. To address overfitting, the model incorporated overlapping of outputs within the pooling layer.

The architecture of the AlexNet model consists of eight layers, with the first five layers being convolutional and the remaining three fully connected. In the original model, the final layer's output was passed through a Softmax function to generate a distribution over 1000 classes. Due to its relatively shallow structure, the AlexNet model generally outperforms more complex CNN models when working with smaller datasets.

The first convolutional layer convolves an image of size 224x224x3 with 96 kernels, each having a size of 11x11x3. The output of the first layer is then filtered by the second layer using 256 kernels of size 5x5x48. The third, fourth, and fifth layers are connected, without normalization or pooling layers, and take the output of the second layer. The third layer consists of 384 kernels of size 3x3x256, the fourth layer also contains 384 kernels of size 3x3x192, and the fifth layer employs 256 kernels of the same size, 3x3x192. Each of the fully connected layers at the end of the model comprises 4096 neurons. Figure 3 provides a visual illustration of the AlexNet model.

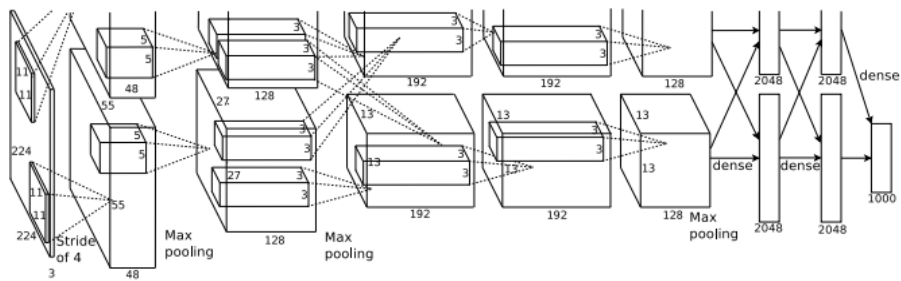

**Fig. 3.** Illustration of the AlexNet model [23].

**Inception-v3 Model.** Following the success of AlexNet in 2012, researchers dedicated their efforts to finding CNN models that could achieve even better performance in computer vision tasks. This pursuit led to the development of the VGGNet [24] and GoogLeNet [25] models in 2014. While these models delivered excellent results, they also had significant drawbacks. VGGNet, although simpler than the Inception architecture of GoogLeNet, required substantial computation time for evaluation. On the other hand, while GoogLeNet had fewer parameters (60 million) compared to VGGNet and AlexNet, its deep complexity posed challenges for scalability. To address these limitations, Szegedy et al. [26] proposed and implemented a novel method to optimize the Inception architecture.

In the study conducted by Szegedy et al. [26], several improvements were made to the Inception architecture. These enhancements included factorizing larger convolutions into multiple convolutions with smaller spatial filters and employing spatial factorization through asymmetric convolutions. Additionally, a notable feature of the Inception architecture was the incorporation of an auxiliary classifier. Deep networks often encounter the vanishing gradient problem, and the auxiliary classifier was designed to mitigate this issue by facilitating the propagation of useful gradients to the lower layers of the network, thereby making them immediately beneficial.

Figure 4 illustrates the improved Inception model, while Figure 5 depicts the inception module after the factorization of n×n convolutions. Figure 6 showcases the inception module with expanded filter bank outputs. An overview of the Inception-v3 architecture, as described by the authors in [26], can be found in Table 2.

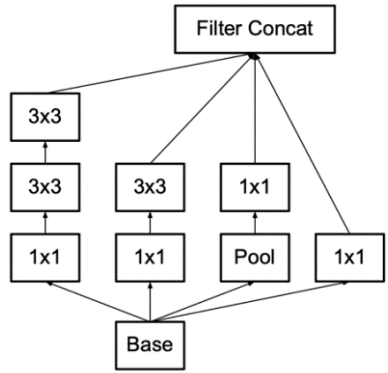

**Fig. 4.** Illustration of the improved Inception module [26]

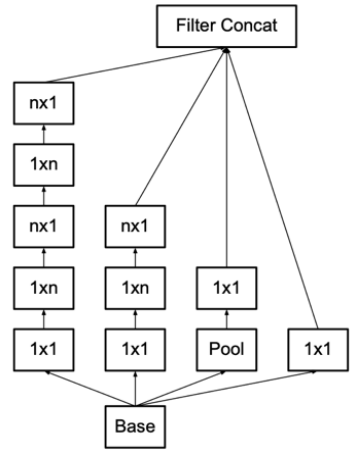

**Fig. 5.** Illustration of the Inception module after factorization of $n \times n$ convolutions [26].

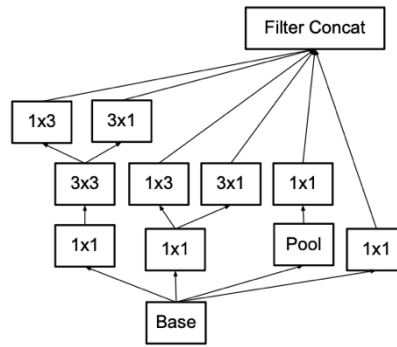

**Fig. 6.** Illustration of the Inception module with expanded filter bank outputs [26]

**Table 2.** Overview of the Inception-v3 Architecture – Input size is the previous layer's output.

| Layer Type | Patch size / stride or remarks | Input size |
|---|---|---|
| Convolutional | 3 x 3 / 2 | *299 x 299 x 3* |
| Convolutional | 3 x 3 / 1 | *149 x 149 x 32* |
| Convolutional Padded | 3 x 3 / 1 | *147 x 147 x 32* |
| Pooling | 3 x 3 / 2 | *147 x 147 x 64* |
| Convolutional | 3 x 3 / 1 | *73 x 73 x 64* |
| Convolutional | 3 x 3 / 2 | *71 x 71 x 80* |
| Convolutional | 3 x 3 / 1 | *35 x 35 x 192* |
| 3 x Inception | Following Figure 4 | *35 x 35 x 288* |
| 5 x Inception | Following Figure 5 | *17 x 17 x 768* |
| 2 x Inception | Following Figure 6 | *8 x 8 x 2048* |
| Pooling | 8 x 8 | *8 x 8 x 2048* |
| Linear | Logits | *1 x 1 x 2048* |
| *Softmax* | *Classifier* | *1 x 1 x 1000* |

**ResNet-18 Model.** Through the works conducted by Szegedy et al. [25, 26] and Simonyan et al. [24], it becomes apparent that deeper CNN models yield better results in image classification. This is attributed to the inherent ability of deeper networks to integrate low, middle, and high-level features with classifiers, thereby creating effective end-to-end models. While Szegedy et al. [26] presented a framework for scaling up networks, such as the Inception architecture that gave rise to the Inception-v3 model, the optimization problem still persists. As the network becomes deeper, the training error rate tends to increase.

To address this challenge, researchers at Microsoft introduced residual learning to the architecture of CNNs, resulting in the creation of the ResNet architecture [27]. Their proposed solution involves utilizing the additional layers of a network for identity mapping, while the remaining layers form a trained shallower network. Shortcut connections, enabling residual learning, are implemented at regular intervals of convolutional layers to facilitate effective identity mapping. The ResNet architecture draws inspiration from the VGG architecture but employs fewer filters. Figure 7 provides an illustration of the ResNet-18 architecture.

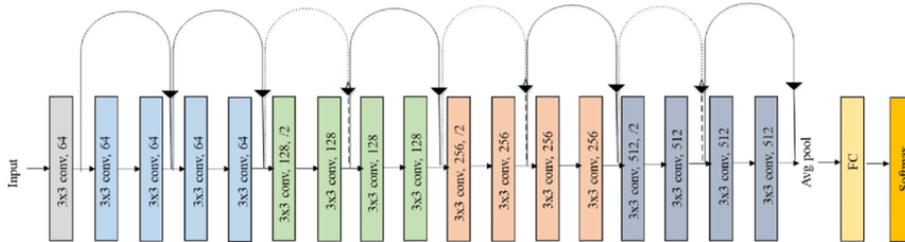

**Fig. 7.** ResNet-18 Architecture [27]

### 3.4 ViT Model

The transformer architecture has emerged as the dominant architecture for Natural Language Processing (NLP) due to its self-attention mechanism. One advantage of this architecture is that it does not suffer from performance degradation as the model scales up, in contrast to CNNs. This was a key issue that motivated researchers at Microsoft to develop the ResNet architecture. However, when it comes to image recognition and classification, utilizing CNNs with self-attention poses challenges in terms of scalability to hardware accelerators. As a result, ResNets continue to be considered the gold standard for CNN-based image recognition and classification tasks.

To address the challenge of self-attention in image classification, researchers at Google drew inspiration from NLP and introduced the Vision Transformer (ViT) architecture [11]. In their work, the image is divided into patches and treated similarly to word tokens in NLP. However, the researchers acknowledged that CNNs possess a stronger inductive bias. Nevertheless, by adopting the standard NLP procedure and fine-tuning a ViT model trained on a significantly larger dataset, this issue has been overcome.

A typical ViT model takes embedded image patches as input, which are then processed by a Transformer encoder. The encoder consists of alternating layers of multi-headed self-attention blocks and multilayer perceptron (MLP) blocks. Each block is connected by a residual connection, and preceding each block is a Layernorm (LN) operation. Following the Transformer encoder is an MLP head for classification. By treating images as word tokens, the ViT architecture avoids introducing inductive bias specific to images. This enables scalability to handle datasets of any size, making ViT a promising approach for image classification in large-scale data scenarios. Figure 8 provides an overview of the ViT architecture as depicted by the authors in [11].

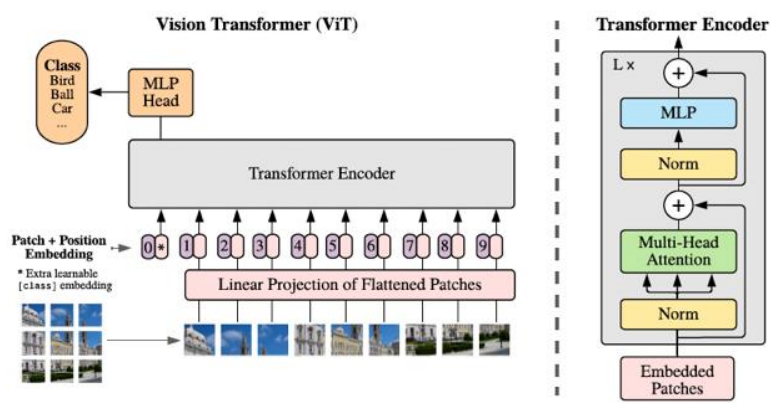

**Fig. 8.** ViT Architecture [11]

### 3.5 Transfer Learning

As previously mentioned, traditional machine learning models required human annotation of features for effective classification. In the medical field, this approach offers no advantages to pathologists as it remains a laborious task, and the quality of features depends on the expertise of the pathologist. To address this, deep learning (DL) models, particularly CNN models, can autonomously extract features from images using their convolutional layers. However, CNNs typically demand large amounts of data, which are currently limited in the medical domain. To overcome this challenge, researchers employ transfer learning, a common practice in NLP.

Transfer learning involves training a model on a large dataset and then fine-tuning it on smaller, specialized data to solve a specific problem. In computer vision tasks, the ImageNet dataset from 2012 has emerged as the primary choice for pretraining DL models. In this research, the DL models were pretrained on the ImageNet dataset and subsequently modified by replacing their classification layers with a fully connected layer that outputs two classes: benign or malignant. This approach allows leveraging the pre-existing knowledge learned from ImageNet to enhance the performance of DL models in classifying medical images.

## 4      Results and Discussion

In this study, an experiment was conducted using each magnification factor of the BreakHis dataset as an independent dataset. For each dataset, 80% of the data was allocated for training and validation, while the remaining 20% was kept separate for final evaluation of the model. The training strategy employed was k-fold cross-validation, with k set to 10. The pre-trained architectures were trained twice: one with full fine-tuning, updating all network weights, and the other by training only the final fully connected layer to assess the model's performance as a fixed feature extractor.

The impact of batch sizes on the generalizability of CNNs for histopathological images was investigated in a study by Kandel et al. [28]. Their findings demonstrated that increasing the batch size does not necessarily improve the generalizability of a CNN model. Moreover, the research indicated that a batch size of 32, coupled with a learning rate of 0.0001 using the Stochastic Gradient Descent (SGD) optimizer, outperformed a model with a batch size of 256 and a learning rate of 0.001. Based on these insights, it was deemed appropriate to utilize a batch size of 16 with a learning rate of 0.0001, employing the SGD optimizer in this research.

The experiments consisted of ten folds, each spanning ten epochs. The model was trained within each fold, followed by validation. Once the validation phase was completed, the model was tested on the unseen data that had been set aside. For each experiment, ten models were created, one for each fold, and the reported results represent the average performance across those models. The evaluation metrics used to obtain the results from these experiments are as follows:

*Accuracy*: This determines the ratio of correctly classified instances over the entire number of instances:

$$Accuracy = \frac{TP + TN}{TP + TN + FP + FN} \times 100$$

*Sensitivity*: This determines the percentage of true positives and it is computed as follows:

$$Sensitivity = \frac{TP}{TP + FN} \times 100$$

*Precision*: Measures the number of times the label of the Positive class has been incorrectly predicted as belonging to another class:

$$Precision = \frac{TP}{TP + FP} \times 100$$

*Specificity*: Measures the proportion of correctly labeled true negatives:
$$Specificity = \frac{TN}{TN + FP} \times 100$$

*F1-score*: This represents the balance between Recall and Precision:
$$F1-Score = \frac{2 \times sensitivity \times precision}{sensitivity + precision} \times 100$$

**Table 3.** Results for breast cancer lesion diagnosis using a fine-tuned AlexNet

| Evaluation | 40x | 100x | 200x | 400x |
|---|---|---|---|---|
| Accuracy % | 81.75 | 82.39 | 90.15 | **81.59** |
| Sensitivity % | 99.14 | 99.24 | 98.2 | **92.8** |
| Precision % | 79.86 | 80.21 | 88.54 | **81.3** |
| Specificity % | 41.17 | 44.21 | 73.88 | **62.79** |
| **F1-Score** | **88.41** | **88.69** | **93.06** | **86.38** |

**Table 4.** Results for breast cancer lesion diagnosis using an AlexNet model with only the classification layer trained.

| Evaluation | 40x | 100x | 200x | 400x |
|---|---|---|---|---|
| Accuracy % | 73.4 | 75.17 | 78.86 | **70.44** |
| Sensitivity % | 99.86 | 99.03 | 97.70 | **96.75** |
| Precision % | 72.56 | 74 | 76.98 | **68.81** |
| Specificity % | 11.67 | 21.09 | 40.90 | **26.32** |
| **F1-Score** | **84.03** | **84.70** | **86.08** | **80.39** |

**Table 5.** Results for breast cancer lesion diagnosis using a fine-tuned Inception-v3

| Evaluation | 40x | 100x | 200x | 400x |
|---|---|---|---|---|
| Accuracy % | 91.56 | 91.34 | 90.99 | **81.48** |
| Sensitivity % | 98.71 | 97.44 | 97.40 | **94.21** |
| Precision % | 90.22 | 90.84 | 90 | **79.87** |
| Specificity % | 75 | 77.5 | 78.06 | **60.15** |
| **F1-Score** | **94.27** | **94** | **93.53** | **86.44** |

**Table 6.** Results for breast cancer lesion diagnosis using an Inception-v3 model with only the classification layer trained.

| Evaluation | 40x | 100x | 200x | 400x |
|---|---|---|---|---|
| Accuracy % | 74.35 | 79.23 | 76.83 | **67.31** |
| Sensitivity % | 99.64 | 99.93 | 99.70 | **98.68** |
| Precision % | 73.33 | 77.05 | 74.41 | **66.08** |
| Specificity % | 15.33 | 32.34 | 30.75 | **14.71** |
| **F1-Score** | **84.48** | **86.99** | **85.21** | **79.12** |

**Table 7.** Results for breast cancer lesion diagnosis using a fine-tuned ResNet-18

| Evaluation | 40x | 100x | 200x | 400x |
|---|---|---|---|---|
| Accuracy % | 89.35 | 82.58 | 85.69 | **74.34** |
| Sensitivity % | 96.93 | 99.31 | 99.56 | **95.26** |
| Precision % | 88.90 | 80.35 | 82.64 | **72.5** |
| Specificity % | 71.67 | 44.69 | 57.76 | **39.26** |
| **F1-Score** | **92.71** | **88.81** | **90.30** | **82.32** |

**Table 8.** Results for breast cancer lesion diagnosis using a ResNet-18 model with only the classification layer trained.

| Evaluation | 40x | 100x | 200x | 400x |
|---|---|---|---|---|
| Accuracy % | 80.15 | 75.89 | 76.28 | **66.42** |
| Sensitivity % | 96.64 | 99.51 | 99.70 | **97.98** |
| Precision % | 79.60 | 74.43 | 74 | **65.56** |
| Specificity % | 41.67 | 22.34 | 29.10 | **13.53** |
| **F1-Score** | **87.23** | **85.15** | **84.92** | **78.53** |

**Table 9.** Results for breast cancer lesion diagnosis using a fine-tuned ViT

| Evaluation | 40x | 100x | 200x | 400x |
|---|---|---|---|---|
| Accuracy % | 92.7 | 91.72 | 95.15 | **88.96** |
| Sensitivity % | 99.5 | 99.17 | 99.26 | **96.58** |
| Precision % | 91.16 | 90.37 | 93.99 | **87.72** |
| Specificity % | 76.83 | 74.84 | 86.87 | **76.18** |
| **F1-Score** | **95.08** | **94.45** | **96.51** | **91.74** |

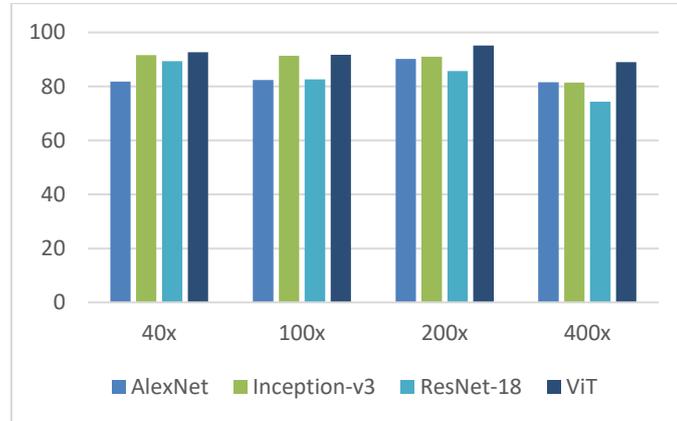

**Fig. 9.** Accuracies of fine-tuned DL models

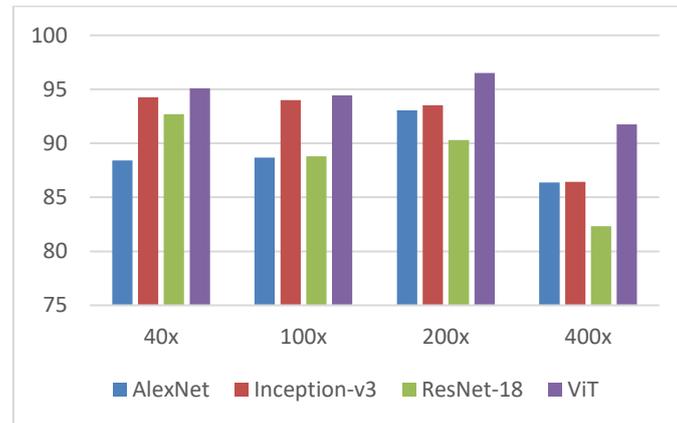

**Fig. 10.** Display of the F1-Scores of fine-tuned DL models

Based on the results presented and depicted in Figures 9 and 10, it is evident that fine-tuning CNN models yield significantly better performance compared to models that only have the final classification layer trained. These findings also support the claims made by He et al. [27], highlighting the positive impact of deeper neural networks on performance. Specifically, the Inception-v3 model demonstrated the best results, achieving a remarkable accuracy of 91.56% on images with 40x magnification. Additionally, the results reinforce the importance of data augmentation and preprocessing. All four fine-tuned Inception-v3 models outperformed the models developed in [8], which lacked data augmentation and preprocessing.

Despite the impressive outcomes, the performance of the Inception-v3 model still falls short compared to the ViT model. The ViT model achieved the highest accuracy of 95.15%, surpassing the state-of-the-art system proposed by Senan et al. [12]. However, it should be noted that the ViT model did not outperform the models created by Thomas et al. [19]. One possible explanation could be the effectiveness of the data preprocessing employed by Thomas et al. in [19], which might have been superior to the approach used in this research. It is worth mentioning that some images were affected by being washed out due to the utilization of default values in the stain normalization method [21], potentially impacting the classification results.

## 5     Conclusion and Future Works

This paper focuses on investigating various deep-learning techniques for the purpose of diagnosing breast cancer using histopathological images. The evaluation was conducted on the BreakHis dataset, where the performance of AlexNet, Inception-v3, and ResNet-18 CNN models was assessed and compared to the newly introduced ViT architecture by Google. The results indicate that the Inception-v3 model, when pre-trained, demonstrated the best performance for end-to-end CNN classification. However, overall, the ViT models exhibited superior performance, surpassing even the state-of-the-art CNN classification models mentioned in [12].

This study also highlights the importance of data preprocessing and augmentation in image classification using CNNs. The Inception-v3 model developed in this paper outperformed the models in [8], which were trained on unprocessed images without any data augmentation techniques applied. The significance of this finding emphasizes the need for careful preprocessing and augmentation steps to enhance the performance of CNN-based classification models.

To expand upon the findings of this paper, future research could focus on creating uniform dataset sizes for experimentation purposes. Additionally, the effect of ViT performance on histopathological images could be explored by evaluating traditional image preprocessing techniques like Histogram Equalization and comparing them against stain normalization technique [21]. Furthermore, a comparison between the performance of hierarchal voting technique applied to an Inception-v3 model and a ViT model could provide valuable insights for further analysis..